\documentclass[twoside,11pt]{article}
 
\usepackage{jmlr2e
}

\usepackage{xspace}
\usepackage{amsmath, amssymb, graphicx, bm, bbm}
\usepackage{xcolor}
\usepackage{wrapfig}
\usepackage{hyperref}
\usepackage{bbm}

\definecolor{NavyBlue}{rgb}{0.000000,0.000000,0.501961}
\hypersetup{
    	colorlinks=true,
	filecolor=NavyBlue
}
    
\newcommand{\la}{$(\lambda)$\xspace}
\def\l{\lambda}
\def\g{\gamma}
\newcommand{\eqn}[1]{\begin{align}#1\end{align}}
\newcommand{\eqns}[1]{\begin{align*}#1\end{align*}}
\newcommand{\eqnl}[1]{\begin{flalign}#1\end{flalign}}

\def\w{{\bf w}}
\def\A{{\bf A}}
\def\b{{\bf b}}
\def\X{{\bf X}}
\def\D{{\bf D}}
\def\I{{\bf I}}
\def\P{{\bf P}}
\def\tr{^\top}
\def\lt{\left}
\def\rt{\right}
\def\paran#1{\lt(#1\rt)}
\def\cb#1{\lt\{#1\rt\}}
\def\sb#1{\lt[#1\rt]}
\def\r{{\bf r}}
\def\hr#1{\href{fakelink}{#1}}

\def\E{{\mathrm{E}}}
\def\x{{\bf x}}
\def\q{{\bf q}}
\def\Real{\mathbb{R}}
\def\St{\mathcal{S}}
\def\Act{\mathcal{A}}
\def\gb{{\bf\gbamma}}
\def\L{{\bf\Lambda}}
\def\Verb#1{\lt\Vert#1\rt\Vert}

\def\bx{\bar{\x}}
\def\bxtp{\bx_{t+1}}

\def\Rtp{R_{t+1}}

\def\a{\alpha}

\def\ln{\zeta} 

\def\deq{\overset{\mathrm{def}}{=\joinrel=}}

\newcommand{\lna}{$(\ln)$\xspace}

\def\e{{\bf e}}
\def\h{{\bf h}}

\def\EM{{\bf E}}
\def\C{{\bf C}}
\def\gb{{\bf g}}
\def\H{{\bf H}}
\def\Left{\emph{left}}
\def\Right{\emph{right}}

\def\hr{}
\newcommand{\half}{\frac{1}{2} } 

\allowdisplaybreaks

\begin{document}
%
\title{Multi-step Off-policy Learning \\
Without Importance Sampling Ratios}
\author{ 
\center \name {
A. Rupam Mahmood\qquad Huizhen Yu \qquad Richard S. Sutton}\\
\vspace{0.5cm}
\addr Reinforcement Learning and Artificial Intelligence Laboratory \\
Department of Computing Science, University of Alberta\\ Edmonton, AB T6G 2E8 Canada \\[0.5cm]
\vspace{0.5cm}
}
\maketitle
\begin{abstract}
To estimate the value functions of policies from exploratory data, most model-free off-policy algorithms rely on importance sampling, where the use of importance sampling ratios often leads to estimates with severe variance. It is thus desirable to learn off-policy without using the ratios. However, such an algorithm does not exist for multi-step learning with function approximation. In this paper, we introduce the first such algorithm based on temporal-difference (TD) learning updates. We show that an explicit use of importance sampling ratios can be eliminated by varying the amount of bootstrapping in TD updates in an action-dependent manner. Our new algorithm achieves stability using a two-timescale gradient-based TD update. A prior algorithm based on lookup table representation called Tree Backup can also be retrieved using action-dependent bootstrapping, becoming a special case of our algorithm. In two challenging off-policy tasks, we demonstrate that our algorithm is stable, effectively avoids the large variance issue, and can perform substantially better than its state-of-the-art counterpart.
\end{abstract}

\section{Introduction}
Off-policy learning constitutes an important class of reinforcement learning problems, where the goal is to learn about a designated target policy while behaving according to a different policy.
In recent years, off-policy learning has garnered a substantial amount of attention in policy evaluation tasks. When safety and data frugality is paramount, the ability to evaluate a policy without actually following it can be invaluable (cf. \hr{Thomas 2015}). Learning a large number of off-policy predictions is also considered important for model learning, options learning (\hr{Sutton et al.\ 1999}), scalable life-long learning (\hr{White, Modayil \& Sutton 2012}), and knowledge representation (\hr{White 2015}). 

A number of computationally scalable algorithms have been proposed that can learn off-policy without requiring to access or estimate the model of the environment and in that sense are \emph{model-free} (\hr{Precup et al.\ 2000, 2001}, \hr{Maei \& Sutton 2010}, \hr{Maei 2011}, \hr{Yu 2012}, \hr{Geist \& Scherrer 2014}, \hr{Dann et al.\ 2014}, \hr{van Hasselt et al.\ 2014}, \hr{Mahmood et al.\ 2014}, \hr{\mbox{Yu 2015}, 2016}, \hr{Hallak et al.\ 2015a}, \hr{Mahmood et al.\ 2015}, \hr{Sutton et al.\ 2016}, \hr{White \& White 2016a}). A core component of these algorithms is a classical Monte Carlo technique called importance sampling (\hr{Hammersley \& Handscomb 1964}, \hr{Rubinstein 1981}), where samples are scaled by the likelihood ratio of the two policies, also known as the \emph{importance sampling \mbox{ratio}}, so that they appear to be drawn from the target policy. Although importance sampling plays a key role in correcting the discrepancy between the policies, the highly varying nature of importance sampling ratios often results in large estimation variance (\hr{Liu 2001}, \hr{Koller \& Friedman 2009}, 
\hr{Dann et al.\ 2014}, 
\hr{Mahmood \& Sutton 2015}, \hr{White \& White 2016a}). Some off-policy algorithms avoid importance sampling using a model of the environment or combine model-based estimation with importance sampling based estimation. Due to performing model estimation in addition to value function estimation, these algorithms tend to be computationally more complex (\hr{Dud\'{i}k et al.\ 2011}, \hr{Paduraru 2013}, 
\hr{Hallak et al.\ 2015b}, 
\hr{Li et al.\ 2015}, 
\hr{Jiang \& Li 2015}, 
\hr{Thomas \& Brunskill 2016}).

Multi-step learning is one of the most important components of modern temporal-difference learning algorithms. Through multi-step learning, temporal-difference learning algorithms can vary smoothly across a large spectrum of algorithms, 
 including both one-step and Monte Carlo methods.
The influence of multi-step learning is the greatest when parametric function approximation is used. In this case, solutions produced by multi-step learning can be much superior to those produced by one-step learning (\hr{Tsitsiklis \& Van Roy 1997}). Unfortunately, the problem of large variance with importance sampling is also the most severe in multi-step learning (\hr{White 2015}, \hr{Mahmood \& Sutton 2015}). Consequently, multi-step off-policy learning remains problematic and largely unfulfilled. 
 
An obvious approach to solve the problem of multi-step off-policy learning would then be to develop an algorithm that avoids using importance sampling ratios. The absence of these ratios will presumably reduce the estimation variance, making long-term multi-step learning tenable. Only a few model-free algorithms have been proposed to learn off-policy without using importance sampling ratios (\hr{Precup et al.\ 2000}, \hr{van Hasselt 2011}, \hr{Harutyunyan et al. 2016}). However, all these algorithms were introduced either for one-step learning (\hr{van Hasselt 2011}) or for learning with lookup table representation (\hr{Precup et al.\ 2000}, \hr{Harutyunyan et al. 2016}), that is, without using parametric function approximation. Multi-step learning does not have a lasting influence on performance in this case. 

Our key contribution is to develop an algorithmic technique based on modulating the amount to which the estimates of the subsequent states are used, a concept known as \emph{bootstrapping}, in an action-dependent manner.
It results in an action-dependent bootstrapping parameter, which is a generalization of the state-dependent bootstrapping parameter used in prior works (\hr{Maei \& Sutton 2010}, \hr{Sutton et al.\ 2014}). 
For action-value estimation, we show that importance sampling ratios can be eliminated by varying the action-dependent bootstrapping parameter for different state-action pairs in a particular way.
Using this technique, we introduce a new algorithm called \emph{ABQ} that can achieve much less estimation variance compared to the state-of-the-art off-policy algorithm. ABQ is the first to effectively achieve multi-step function approximation solutions for off-policy learning without explicitly using importance sampling ratios. A prior algorithm, Tree Backup (\hr{Precup et al.\ 2000}), can be retrieved as a special case of our algorithm. Furthermore, we show that \mbox{another} off-policy algorithm, Retrace (\hr{Munos et al.\ 2016}), can also be derived and extended to the case of function approximation with stability using the action-dependent bootstrapping technique.

\section{Problem formulation and notations}

In this section, we formulate the problem of multi-step off-policy learning with parametric function approximation and establish notations. Consider an agent in a dynamical environment with a finite state space $\St$ and action space $\Act$. 
At each time $t=0, 1, \ldots$, if the present state is $s \in\St$ and the agent takes action $a \in\Act$, then the next state $S_{t+1}$ is $s' \in \St$ with probability $p(s' |s,a)$, and the agent receives a random reward $R_{t+1}$ with mean $r(s,a)$ and finite variance upon the state transition.
A randomized stationary policy $\pi$ specifies the probability $\pi(a|s)$ of taking action $a$ at state $s$.
Of our interest is a given policy $\pi$, referred to as the target policy, and the performance of the agent if it follows $\pi$. 
Specifically, our interest in this paper is to estimate the action-value function of $\pi$, defined as the expected sum of discounted rewards for any initial state-action pair $(s,a)$:
\eqn{
q_\pi(s,a) \deq \E_\pi\sb{ \sum_{t=1}^{\infty} \g^{t-1} R_{t} \Big| S_0=s, A_0=a },
}
where $\g<1$ is a discount factor, and $\E_\pi\sb{\cdot}$ signifies that all subsequent actions follow \mbox{policy $\pi$.} 

In off-policy learning, the goal is to estimate $q_\pi$ from the observations $\{(S_t, A_t, R_{t+1})\}_{t \geq 0}$ obtained by the agent while it follows a (stationary randomized) policy $\mu \not= \pi$. We call $\mu$ the behavior policy. (The case $\mu=\pi$ is the on-policy case.) 
We shall assume that $\mu$ induces an irreducible Markov chain on the state-action space, whose unique invariant distribution we denote by $d_\mu$.

Important to multi-step learning are multi-step Bellman equations satisfied by the action-value function $q_\pi$. 
We review here such equations for the well-known TD($\lambda$), where $\l \in [0,1]$ is the bootstrapping parameter. Let $\P_\pi$ be the transition probability matrix of the Markov chain on $\St \times \Act$ induced by the target policy $\pi$, and let $\r\in\Real^{|\St|\cdot|\Act|}$ be the vector of expected rewards for different state-action pairs: $[\r]_{sa} \deq r(s,a)$.\footnote{We use brackets with subscripts to denote elements of vectors and matrices. We use small letters for vectors and capital letters for matrices, both boldfaced.} 
For $\l \in [0,1]$, define the multi-step Bellman operator $T_\pi^{(\l)}$ by
\eqns{
T_\pi^{(\l)}\q \deq \paran{\I - \g\l\P_\pi}^{-1} \!\!\sb{\r + \g(1-\l)\P_\pi\q}\! 
}
for all $\q\!\in\!\Real^{|\St|\cdot|\Act|}$, where $\I$ is the identity matrix. Then $q_\pi$ satisfies the multi-step Bellman equation $\q_\pi = T_\pi^{(\l)} \q_\pi$, where $\q_\pi$ stands for the action-value function in vector notation. 

We approximate the action-value function as a linear function of some given features of state-action pairs: $q_\pi(s,a) \approx \w\tr \x(s,a)$, where $\w\in\Real^n$ is the parameter vector to be estimated and $\x(s,a) \in\Real^n$ is the feature vector for state $s$ and action $a$. 
In matrix notation we write this approximation as $\q_\pi \approx \X\w$, where $\X\in\Real^{|\St|\cdot|\Act|\times n}$ is the feature matrix with the rows being the feature vectors for different state-action pairs: $[\X]_{sa,:} = \x(s,a)\tr$. 

The multi-step solution to the off-policy learning problem with function approximation can be found by solving the fixed point equation: $\X\w = \Pi_\mu T_\pi^{(\l)} \X\w$ (when it has a solution), where $\Pi_\mu \deq \X(\X\tr \D_\mu \X)^{-1}\X\tr \D_\mu$ is the projection matrix with $\D_\mu \in \Real^{|\St|\cdot|\Act|\times|\St|\cdot|\Act|}$ being a diagonal matrix with diagonal elements $d_\mu(s,a)$. The Mean Squared Projected Bellman Error (MSPBE) corresponding to this equation is given by:
\eqn{
J(\w)		&= \Verb{ \Pi_\mu T_\pi^{(\l)}\X\w - \X\w  }_{\D_\mu}^2,
}
where $\Verb{\q}^2_{\D_\mu} = \q\tr \D_\mu \q$, for any $\q\in\Real^{|\St|\cdot|\Act|}$.
The multi-step asymptotic TD solution associated with the fixed-point equation and the above MSPBE can be expressed as $\w_\infty \deq \A^{-1}\b$, when $\A$ is an invertible matrix, and $\A$ and $\b$ are given by
\eqn{
&\A 	\deq \X\tr \D_\mu \paran{\I - \g \l \P_\pi }^{-1} \paran{\I-\g \P_\pi} \X,& \label{eqn:A}\\
&\b \deq \X\tr \D_\mu \paran{\I - \g \l \P_\pi }^{-1} \r. & \label{eqn:b}
}

\section{The advantage of multi-step learning}

Under the rubric of temporal-difference learning fall a broad spectrum of methods. 
On one end of the spectrum, we have one-step methods that fully bootstrap using estimates of the next state and use only the immediate rewards as samples. On the other end of the spectrum, we have Monte Carlo methods that do not bootstrap and rather use all future rewards for making updates. Many multi-step learning algorithms incorporate this full spectrum and can vary smoothly between one-step and Monte Carlo updates using the {bootstrapping parameter} $\lambda$. Here $1-\lambda$ determines the degree to which bootstrapping is used in the algorithm. With $\lambda=0$, these algorithms achieve one-step TD updates, whereas with $\lambda=1$, they effectively achieve Monte Carlo updates. To contrast with one-step learning, multi-step learning is generally viewed as learning with $\lambda>0$ in TD methods.

Multi-step learning impacts the efficiency of estimation in two ways. First, it allows more efficient estimation compared to one-step learning with a finite amount of samples. One-step learning uses the minimal amount of samples, have relatively less variance compared to Monte Carlo updates, but produces biased estimates. Typically, with a finite amount of samples, a value of $\lambda$ between 0 and 1 reduces the estimation error the most (\hr{Sutton \& Barto 1998}).

\begin{figure}[h]
\begin{center}
\hspace{-10pt}
\includegraphics[scale=1.3]{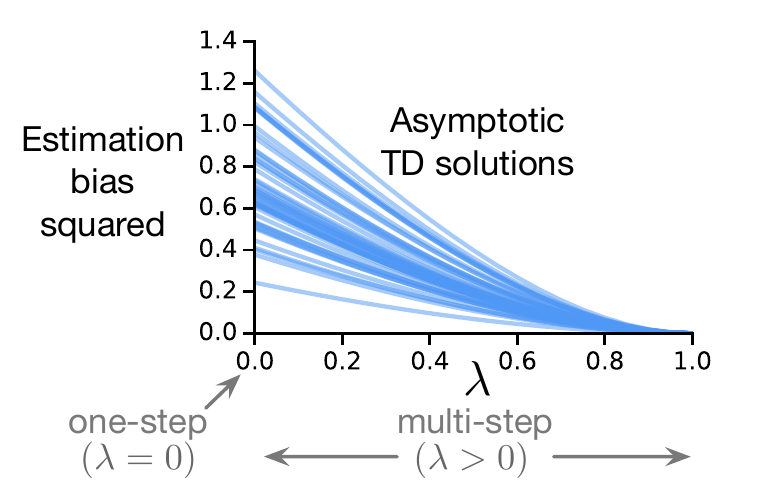}
\caption{
Multi-step solutions are generally superior to one-step solutions, as estimation bias goes to zero often monotonically with increasing $\lambda$, shown here for 50 randomly constructed MDPs. In these MDPs, we used 100 states, 5 actions, and 40 features. The rewards, probabilities, and feature values (binary) were chosen uniformly randomly.}
\label{fig:asymp-bias}
 \end{center}
\end{figure}

Second, when function approximation is used and $q_\pi$ does not lie in the approximation subspace, multi-step learning can produce superior  asymptotic solutions compared to one-step learning.
As $\l$ increases, the multi-step Bellman operator approaches the constant operator that maps every $q$ to $q_\pi$. This in general leads to better approximations, as suggested by the monotonically improving error bound of asymptotic solutions (\hr{Tsitsiklis \& Van Roy 1997}) in the on-policy case, and as we demonstrate for the off-policy case in Figure \ref{fig:asymp-bias}.


Although multi-step learning is desirable with function approximation, it is more difficult in the off-policy case where the detrimental effect of importance sampling is most pronounced. For this reason, off-policy learning without importance sampling ratios is a naturally appealing and desirable solution to this problem. Prior works on off-policy learning without the ratios (e.g., \hr{Precup et al.\ 2000}, \hr{Harutyunyan et al.\ 2016}) are given in the lookup table case where the benefit of multi-step learning does not show up, because regardless of $\l$, the asymptotic solution is $q_\pi$. It is in the case of function approximation that multi-step off-policy learning without importance sampling ratios is most needed.

\section{Multi-step off-policy learning with importance sampling ratios}

To set the stage for our work, we describe in this section the canonical multi-step off-policy learning update with importance sampling ratios, and how the ratios introduce variance in off-policy temporal-difference (TD) updates. A TD update is generally constructed based on stochastic approximation methods, where the target of the update is based on returns. Here we consider the off-line update for off-policy TD learning. Although not practical for implementation, off-line updates are useful and develop the foundation for deriving practical and computationally efficient algorithms (Sutton \& Barto 1998, Seijen et al. 2016).
An off-line TD update for off-policy action-value estimation based on multi-step returns can be defined as:
\eqn{
\Delta\w_{t} = \alpha_t \lt( G^{\l}_t - \w\tr\x_t \rt) \x_t, \label{eqn:off-policy-q-lambda}
}
where $\a>0$ is the step-size parameter, and $\w$ is a fixed weight vector.  Here, $G^{\l}_t$ is the multi-step target, known as $\l$-return, defined as the sum of TD errors weighted by powers of $\g\l$ and products of importance sampling ratios:
\eqn{
G^{\l}_t &\deq  \sum_{n=t}^\infty  (\g\l)^{n-t} \rho_{t+1}^{n} \delta_{n}  + \w\tr\x_t.
\label{eqn:l-return}
}
The TD error $\delta_t$ is defined as $\delta_t\deq \Rtp + \gamma\w\tr\bxtp - \w\tr\x_t$, with $\bx_t \deq \sum_{a} \pi(a|S_t) \x(S_t,a)$.
The term $\rho_t^n \deq \Pi_{i=t}^{n} \rho_i$ is a product of importance sampling ratios $\rho_t\deq \frac{ \pi(A_t|S_t) }{\mu(A_t|S_t)}$, which accounts for the discrepancy due to using the behavior policy instead of the target policy throughout the trajectory.
 Note that, the update defined by (\ref{eqn:off-policy-q-lambda}) is a forward-view update, that is, it uses samples that only become available in the future from the time the state of the updated estimate is visited.
We call this update \emph{the off-policy Q\la update}. It can be shown that the asymptotic multi-step solution corresponding to off-policy Q\la is given by $\w_\infty = \A^{-1}\b$, (\ref{eqn:A}), and (\ref{eqn:b}), when $\A$ is invertible. All existing multi-step off-policy algorithms with importance sampling are of this form or a variant.

When $\l=0$, no importance sampling ratios are involved, and this update reduces to that of off-policy expected Sarsa (\hr{Sutton \& Barto 1998}, \hr{van Hasselt 2011}, \hr{Sutton et al.\ 2014}). 
This one-step update is also closely related to the one-step Q-learning update, where a greedy nonstationary target policy is used instead of a fixed stationary one.

The importance sampling ratios play a role when $\l>0$, and their influence, including the detrimental impact on variance, is greater with larger $\l$. The product $\rho_{1}^{n}$ of off-policy Q\la in (\ref{eqn:l-return}) can become as large as $\frac{1}{\lt(\min_{s,a} \mu(a|s)\rt)^{n}}$. 
Such an exponential growth, when occurred even momentarily, can have large impact on the variance of the estimate.
If the value of $\l$ is small or very close to zero, the large variance ensuing from the product may be avoided, but it would also be devoid of much of the benefits of multi-step learning.  

\section{Avoiding importance sampling ratios}
In this section, we introduce the idea of action-dependent bootstrapping and how it can be used to avoid importance sampling ratios in off-policy estimates. For that, first we introduce an action-dependent bootstrapping parameter $\l(s,a)\in[0,1]$, which is allowed to vary between different state-action pairs. A closely related idea is state-dependent bootstrapping used by Sutton and Singh (\hr{1994}) and Sutton et al. (\hr{2014}) for state-value estimation, and by Maei and Sutton (\hr{2010}) for action-value estimation. In those works, the degree of bootstrapping was allowed to vary from one state to another by the state-dependent bootstrapping parameter $\l(s)\in[0,1]$ but was not used as a device to reduce the estimation variance.

The variability of the parameter $\l(s,a)$ can be utilized algorithmically on a moment-by-moment basis to absorb the detrimental effect of importance sampling and in general to control the impact of importance sampling. 
Let us use the notational shorthand $\l_t\deq\l(S_t, A_t)$, and define a new $\l$-return by replacing the constant $\l$ in (\ref{eqn:l-return}) with variable $\l(s,a)$:
\eqn{
G^{\l}_t &= \sum_{n=t}^\infty  \g^{n-t} \l_{t+1}^{n} \rho_{t+1}^{n} \delta_{n}  + \w\tr\x_t,
\label{eqn:l-return-two}
}
where $\l_t^n \deq \Pi_{i=t}^{n} \l_i$.
Notice that each importance sampling ratio in (\ref{eqn:l-return-two}) is factored with a corresponding bootstrapping parameter: $\l_t \rho_t$. 
We can mitigate an explicit use of importance sampling ratios by setting the action-dependent bootstrapping parameter $\l(s,a)$ in the following way:
\eqn{
\l(s,a) = \nu(\psi, s, a) \mu(a|s), ~~~~~~~~\nu(\psi, s, a)\deq\min\paran{\psi, \frac{1}{\max\paran{\mu(a|s), \pi(a|s) }}}
}
where $\psi\ge0$ is a constant.
Note that $\nu(\psi, s, a)$ is upper-bounded by $\psi_{\max}$, which is defined as follows:
\eqn{\psi_{\max} \deq  \frac{1}{\min_{s,a}\max\paran{\mu(a|s), \pi(a|s) }}.
} 
The product $\l_t\rho_t$ can then be rewritten as: $\l_{t}\rho_t = \nu(\psi, S_t, A_t) \mu_t \frac{\pi_t}{\mu_t} = \nu(\psi, S_t, A_t)\pi_t$, dispelling an explicit presence of importance sampling ratios from the update. 
It is easy to see that, under our proposed scheme, the effective bootstrapping parameter is upper bounded by 1: $\l_{t} \le 1$, and at the same time all the products are also upper bounded by one: ${\l}_{t}^{n} \rho_{t}^{n}\le 1$, largely reducing variance.

\begin{figure}[t]
\centering
\hspace{-0pt}
	\includegraphics[scale=0.9]{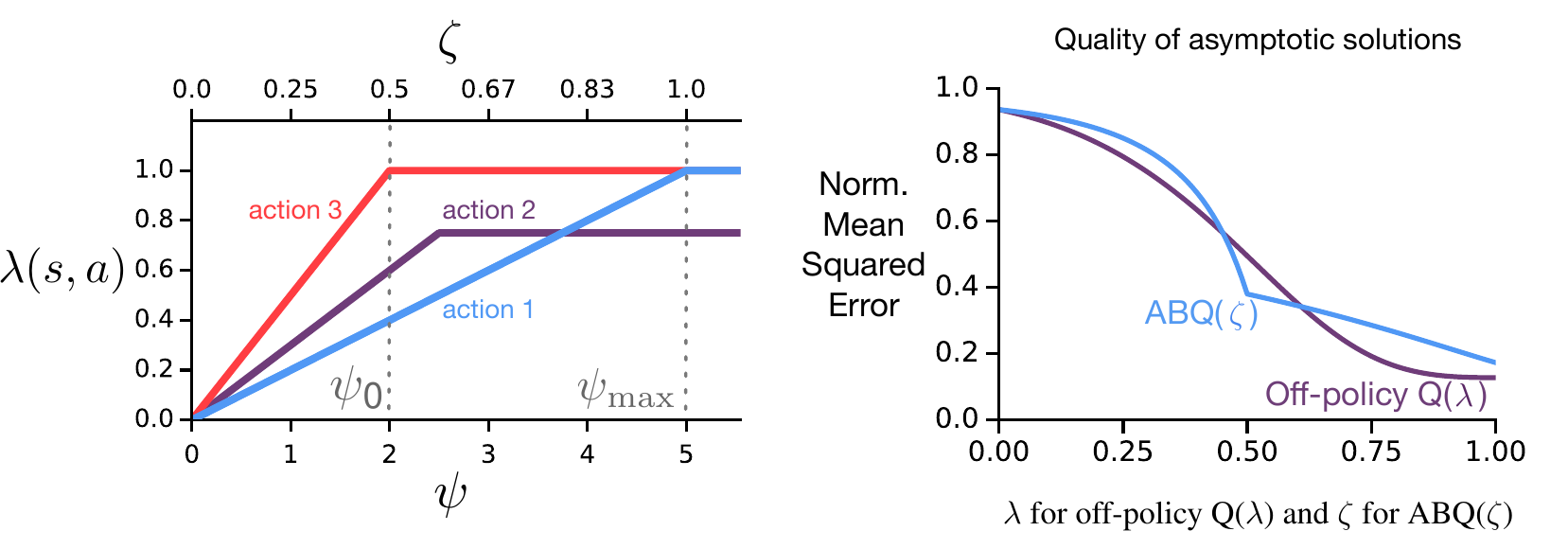}
\caption{ {\bf Left}: The effect of $\psi$ and $\ln$ on $\l(s,a)$ for three different actions under the action-dependent bootstrapping scheme.
As $\psi$ is increased, the parameter $\l$ for different actions increase at a different rate and get capped for different values of $\psi$. \newline
{\bf Right}: 
The effect of $\l$ on GQ\la solutions and $\ln$ on ABQ\lna solutions. \!Multi-step ($\l>0$) off-policy Q\la solutions are superior to the one-step ($\l=0$) solution. ABQ\lna solutions can also achieve a similar multi-step advantage with $\ln>0$. }
\label{fig:zeta-tuning-and-asymp-sol}
\end{figure}

To understand how $\psi$ influences $\l(s,a)$ let us use the following example, where there are only one state and three actions $\cb{1, 2, 3}$ available. The behavior policy probabilities are $\sb{0.2, 0.3, 0.5}$ and the target policy probabilities are $\sb{0.2, 0.4, 0.4}$ for the three actions, respectively. Figure \ref{fig:zeta-tuning-and-asymp-sol} (left) shows how the action-dependent bootstrapping parameter $\l$ for different actions change as $\psi$ is increased from 0 to $\psi_{\max}$. Initially, the bootstrapping parameter $\l$ increased linearly for all actions at a different rate depending on their corresponding behavior policy probabilities. The $\min$ in the factor $\nu$ comes into effect with $\psi > \psi_0 \deq \frac{1}{\max_{s,a} \max\paran{\mu(a|s), \pi(a|s)}}$, and the largest $\l$ at $\psi=\psi_0$ gets capped first. Eventually, with large enough $\psi$, that is, $\psi\ge \psi_{\max}$, all $\l$s get capped.


Algorithms with constant $\l$ are typically studied in terms of their parameters by varying $\l$ between $[0,1]$, which would not be possible for an algorithm based on the above scheme as $\l$ is not a constant any more. For our scheme, the constant tunable parameter is $\psi$, which has three pivotal values: $[0, \psi_0, \psi_{\max}]$. For parameter studies, it would be convenient if $\psi$ is scaled to another tunable parameter between $[0, 1]$. But in that case, we have to make a decision on what value $\psi_0$ should transform to. In the absence of a clear sense of it, a default choice would be to transform $\psi_0$ to 0.5. One such scaling is where we set $\psi$ as a function of another constant $\ln\ge0$ in the following way:
\eqn{
\psi(\ln)=2 \ln \psi_0 + \paran{2 \ln - 1}^+\times\paran{\psi_{\max} - 2 \psi_0 }, \label{eqn:psi-to-zeta}
}
and then vary $\ln$ between $[0, 1]$ as one would vary $\l$. Here $\paran{x}^+ = \max\paran{0, x}$. In this case, $\ln=0, 0.5$, and $1$ correspond to $\psi(\ln)=0, \psi_0$, and $\psi_{\max}$, respectively. 
Note that $\ln$ can be written conversely in terms of $\psi$ as follows:
\eqn{
\ln = \paran{\psi - \psi_0}^+ \cdot \frac{ 2 \psi_0 - \psi_{\max} }{ 2 \paran{ \psi_{\max} - \psi_0 } \psi_0 } + \frac{ \psi }{ 2 \psi_0 }. \label{eqn:zeta-to-psi}}
The top margin of Figure \ref{fig:zeta-tuning-and-asymp-sol} (left) shows an alternate x-axis in terms of $\ln$. 

To form an update using this proposed modification to $\l$, let us use the following notational shorthands, 
\eqn{
&\nu_\ln(s, a)\deq\nu(\psi(\ln), s, a), & \nu_{\ln,t} \deq \nu_{\ln}(S_t, A_t), \label{eqn:nu-shorthand} \\
&\l_\ln(s,a)\deq \nu(\psi(\ln), s, a) \mu(a|s), & \l_{\ln, t}\deq\l_{\ln}(S_t, A_t). \label{eqn:lambda-shorthand}
}
 We use $H_t^\ln$ to denote the $\l$-return defined by (\ref{eqn:l-return-two}) with the bootstrapping parameter set according to $\l_\ln$:
\eqn{
H^{\zeta}_t &= \sum_{n=t}^\infty  \g^{n-t} \l^n_{\zeta,t+1} \rho_{t+1}^{n} \delta_{n}  + \w\tr\x_t = \sum_{n=t}^\infty  \g^{n-t} \nu^n_{\zeta,t+1} \pi_{t+1}^{n} \delta_{n}  + \w\tr\x_t,
\label{eqn:new-l-return}
}
where $\l_{\zeta,t}^n \deq \Pi_{i=t}^{n} \l_{\zeta,i}$, $\nu_{\zeta,t}^n \deq \Pi_{i=t}^{n} \nu_{\zeta,i}$, and $\pi_t^n \deq \Pi_{i=t}^{n} \pi_i$. The off-line forward-view update with $H_t^\ln$ as the target can be written as:
\eqn{
\Delta\w_t = \a_t \paran{ H_t^{\ln} - \w^\top \x_t }\x_t. \label{eqn:abq-for}
}
The asymptotic solution corresponding to this update, which we call the \emph{ABQ\lna} solution, is \mbox{$\w_\infty^\ln \deq {\A_\ln}^{-1}\b_\ln$} with
\eqn{
\A_\ln	&\deq \X\tr \D_\mu \paran{\I - \g \P_\pi  \L_\zeta }^{-1} \paran{\I- \g \P_\pi } \X, \label{eqn:Aabq}\\
\b_\ln	&\deq \X\tr \D_\mu \paran{\I - \g \P_\pi \L_\zeta }^{-1} \r,  \label{eqn:babq}
}
assuming $\A_\zeta$ is invertible.
The derivation is given in Appendix A.1. Here $\L_\zeta$ is a diagonal matrix with $\l_\ln(s,a)$ being the diagonal elements. This is a multi-step solution when the bootstrapping parameter $\l_\ln(s,a)$ does not uniformly reduce to zero. The drawback of this scheme is that we cannot achieve $\l_\ln(s,a)=1$ for all state-action pairs, which would produce the off-policy Monte Carlo solution.  It is the cost of avoiding importance sampling ratios together with its large variance issue. However, much of the multi-step benefits can still be retained by choosing a large $\ln$. 


To illustrate that ABQ\lna solutions can retain much of the multi-step benefits, we used a two-state off-policy task similar to the off-policy task by Sutton et al.\ (\hr{2016}). In this task, there were two states each with two actions, $\Left$ and $\Right$, leading to one of the two states deterministically. More specifically, $p(1|1, \Left)=p(2|1, \Right)=p(1|2, \Left)=p(2|2, \Right)=1$.
There is a deterministic nonzero reward $r(2,\Right)=+1$; all other transitions have reward zero.
The discount factor $\g$ was 0.9. The feature vectors were set as $\x(1,\emph{left})=\x(1,\emph{right})=1$ and $\x(2,\emph{left})=\x(2,\emph{right})=2$. The behavior policy was chosen as $\mu(\emph{right}|1)=\mu(\emph{left}|2)=0.9$ to break away from the uniform distribution of the original problem. The target policy was chosen as $\pi(\emph{right}|\cdot)=0.9$. 

We produced different asymptotic solutions defined by (\ref{eqn:Aabq}) and (\ref{eqn:babq}), choosing different constant $\ln$ between $[0, 1]$. We compared ABQ\lna solutions with off-policy Q\la solutions in terms of the Mean Squared Error (MSE) $\Verb{\X\w-\q_\pi}^2_{\D_\mu}$ normalized by $\Verb{\q_\pi}^2_{\D_\mu}$. The results are given in Figure \ref{fig:zeta-tuning-and-asymp-sol} (right). Off-policy Q\la solutions with $\l>0$ in this task are substantially better than the one-step solution produced with $\l=0$. The ABQ\lna solutions cannot be as good as the off-policy Q(1) solution, as we already anticipated, but much of the benefits of multi-step off-policy solutions can be attained by choosing a large value of $\ln$. Although the tunable parameter $\ln$ of ABQ was set to be a constant, the effective bootstrapping parameter $\l_\ln(s,a)$ was different for different state-action pairs. Therefore, ABQ solutions cannot be obtained simply by rescaling the constant $\l$ of off-policy Q\la.

The algorithm in principle can be used with either $\ln$ or $\psi$ as the tunable parameter. If we tune with $\ln\in[0,1]$, we will have to first use (\ref{eqn:psi-to-zeta}) to obtain $\psi(\ln)$, and then (\ref{eqn:nu-shorthand}) and (\ref{eqn:lambda-shorthand}) to obtain $\l_t$ on a moment-by-moment basis. Computing $\psi(\ln)$ from $\ln$ requires knowing $\psi_0$ and $\psi_{\max}$ which are often known, for example, when the policies are in $\epsilon$-greedy form or fixed and known beforehand. In other cases, tuning with $\psi$ directly can be more convenient. As the scaled parameter $\zeta$ reflects more clearly the qualitative behavior of ABQ, we use it as the tunable parameter in this work.

\section{The ABQ\lna algorithm with gradient correction and scalable updates}

In this section, we develop a computationally scalable and stable algorithm corresponding to the ABQ\lna solution. 
The update (\ref{eqn:abq-for}) given earlier cannot be computed in a scalable manner, because forward-view updates require an increasing amount of computation as time progresses. Moreover, off-policy algorithms with bootstrapping and function approximation may be unstable (Sutton \& Barto 1998), unless machinery for ensuring stability is incorporated. 
Our goal is to develop a stable update corresponding to ABQ\lna solutions while keeping it free from an explicit use of importance sampling ratios.

First, we produce the equivalent backward view of (\ref{eqn:abq-for}) so that the updates can be implemented without forming explicitly the $\l$-return. As we derive in Appendix A.2, these updates are given by
\eqn{
\Delta\w_{t} 	&= \a_t\delta_t \e_t, \qquad \e_t 		=\g \nu_{\zeta,t} \pi_t \e_{t-1} + \x_t. \label{eqn:abq-back}
}
Here, $\e_t\in\Real^n$ is an accumulating trace vector. The above backward-view update achieves equivalence with the forward-view of (\ref{eqn:abq-for}) only for off-line updating, which is typical for all algorithms with accumulating traces. Equivalence for online updating could also be achieved by following the approach taken by van Seijen et al.\ (2014, \hr{2016}), but we leave that out for simplicity.

We take the approach proposed by Maei (\hr{2011}) to develop a stable gradient-based TD algorithm. The key step in deriving a gradient-based TD algorithm is to formulate an associated Mean Squared Projected Bellman Error (MSPBE) and produce its gradient, which can then be sampled to produce stochastic updates.

The MSPBE for the update (\ref{eqn:abq-for}) is given by:
\eqn{
J(\w)		&= \Verb{ \Pi_\mu T_\pi^{(\L_{\zeta})}\X\w - \X\w  }_{\D_\mu}^2  \label{eqn:mspbe}\\
&= \paran{ \X\tr\D_\mu\paran{T_\pi^{(\L_{\zeta})} \X\w - \X\w}}\tr  (\X\tr \D_\mu \X)^{-1} \X\tr \D_\mu\paran{T_\pi^{(\L_{\zeta})}  \X\w - \X\w}.
}
Here, the Bellman operator corresponding to the bootstrapping matrix $\L_\zeta$ is defined for all $\q\!\in\!\Real^{|\St|\cdot|\Act|}$ as 
\[T_\pi^{(\L_{\zeta})} \q \deq \paran{\I - \g\P_\pi\L_{\zeta}}^{-1} \sb{\r + \g\P_\pi(\I-\L_{\zeta}) \q}.\]
Then the gradient can be written as:
\eqn{
\nabla J(\w)		&= -\half\paran{\X\tr\D_\mu{ {  \paran{\I - \g\P_\pi\L_{\zeta}}^{-1}\paran{ \I-\g\P_\pi} \X  } }}\tr \C^{-1} \gb \\
&= - \half\paran{ \gb - \g \H\tr \C^{-1}\gb },
}
where $\gb$, $\C$ and $\H$ are defined in the following way:
\eqn{
\gb&\deq \X\tr\D_\mu\paran{T_\pi^{(\L_{\zeta})} \X\w -  \X\w}, \\
\C &\deq (\X\tr \D_\mu \X), \\ 
\H &\deq\paran{\X\tr\D_\mu\paran{\I - \g  \P_\pi\L_{\zeta}}^{-1} \P_\pi(\I-\L_{\zeta})\X}.
} 
When this gradient is known, the gradient-descent step would be to add the following to the parameter vector:
\eqn{
 -\half \a_t \nabla J(\w)  &=  \a_t \paran{ \gb - \g \H\tr \C^{-1}\gb }.
}
In model-free learning, the gradient is not available to the learner. However, we can form a stochastic gradient descent update by sampling from this gradient. For that, we can express $\gb$ and $\H$ in an expectation form, and estimate the vector $\C^{-1}\gb$ at a faster time scale. 

In Appendix A.3, we derive the resulting stochastic gradient corrected algorithm in details, which we call \emph{the ABQ\lna algorithm}. This algorithm is defined by the following online updates:
{
\eqn{
\delta_t	&\deq \Rtp + \gamma\w_t\tr\bxtp - \w_t\tr\x_t, &\\
\tilde{\x}_{t+1}	&\deq \sum_a \nu_\zeta(S_{t+1}, a) \pi(a|S_{t+1}) \x(S_{t+1}, a), &\\
\e_t 		&\deq \g \nu_{\ln,t} \pi_t \e_{t-1} + \x_t, &\\
\w_{t+1} 	&\!\deq\! \w_t \!+ \a_t\!\!\paran{\!\delta_{t} \e_t \!-\! \g \e_t \tr \h_t \paran{\bxtp - \tilde{\x}_{t+1} }\!}\!, \label{eqn:abq-w} &\\
\h_{t+1}	&\deq \h_t + \beta_t \paran{ \delta_t \e_t - \paran{\h_t\tr\x_t}\x_t }.& \label{eqn:abq-h}
}
}The iteration (\ref{eqn:abq-w}) carries out stochastic gradient-descent steps to minimize the MSPBE (\ref{eqn:mspbe}), and it differs from (\ref{eqn:abq-back}) as it includes a gradient-correction term $-\g \e_t \tr \h_t \paran{\bxtp - \tilde{\x}_{t+1}}$. 
This correction term involves an extra vector parameter $\h_t$. Note that no importance sampling ratios are needed in the update of $\h_t$ or in the gradient-correction term. 
The vector $\h_t$ is updated according to (\ref{eqn:abq-h}) at a faster timescale than $\w_t$ by using a second step-size parameter $\beta_t \gg \alpha_t$ when both step sizes are diminishing. One can achieve that by letting $\alpha_t = O(1/t), \beta_t = O(1/t^c), c \in (1/2, 1)$, for instance. In practice, when stability is not a problem, smaller values of $\beta$ often lead to better performance (\hr{White \& White 2016a}). 




\section{Experimental results}

We empirically evaluate ABQ\lna on three policy evaluation tasks: the two-state off-policy task from Section 5, an off-policy policy evaluation adaptation of the Mountain Car domain (Sutton \& Barto 1998), and the 7-star Baird's counterexample (\hr{Baird 1995}, \hr{White 2015}). In the first two tasks we investigated whether ABQ\lna can produce correct estimates with less variance compared to GQ\la (\hr{Maei 2011}), the state-of-the-art importance sampling based algorithm for action-value estimation with function approximation. We validate the stability of ABQ\lna in the final task, where off-policy algorithms without a stability guarantee (e.g., off-policy Q\la) tend to diverge.

Although the MDP involved in the two-state task is small, off-policy algorithms may suffer severely in this task as the importance sampling ratio, once in a while, can be as large as 9. We simulated both GQ\la and ABQ\la on this task for 10000 time steps, starting with $\w_0={\bf 0}$. We averaged the MSE $\Verb{\X\w_t-\q_\pi}^2_{\D_\mu}$ for the last 5000 time steps. We further averaged this quantity over 100 independent runs. Finally, we divided this error by $\Verb{\q_\pi}^2_{\D_\mu}$ to obtain the normalized MSE (NMSE).

\begin{figure}[t]
\centering
\hspace{-10pt}
	\includegraphics[scale=0.85]{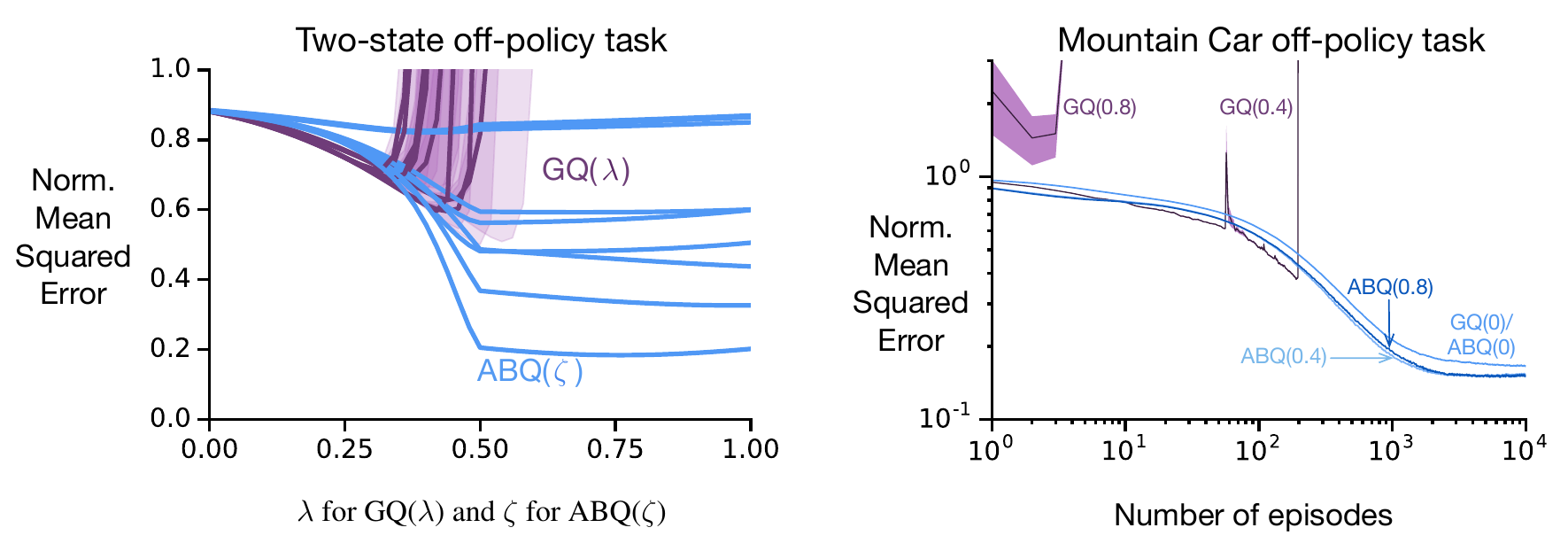}
\caption{ 
{\bf Left}: Comparison of empirical performance of GQ\la and ABQ\lna on a two-state off-policy policy evaluation task. Performance is shown in normalized mean squared error with respect to different values of $\l$ for GQ\la and $\ln$ for ABQ\lna. Different curves are for different combinations of step-size values.
GQ\la produces large MSE when large $\l$ is used. ABQ\lna tolerates larger values of $\zeta$ and thus can better retain the benefits of multi-step learning compared to GQ\la. \newline
{\bf Right}: 
Comparison of empirical performance of GQ\la and ABQ\lna on an off-policy policy evaluation task based on the Mountain Car domain. Each curve shows how learning progresses in terms of estimated normalized mean squared error as more episodes are observed. Different learning curves are for different values of $\l$ for GQ\la and $\ln$ for ABQ\lna. All of them are shown for a particular combination of step-size values. The spikes in GQ\la's learning curves are the result of the occasional large values of importance sampling ratios, increasing the variance of the estimate for GQ\la as large values of $\l$ are chosen. ABQ\lna, on the other hand, can achieve lower mean squared error with larger values of $\ln$ by reducing the estimation variance.
}
\label{fig:two-state-and-mountaincar}
\end{figure}

Figure \ref{fig:two-state-and-mountaincar} (left) shows curves for different combinations of the step-size parameters: $\a\in[0.001, 0.005,0.01]$ and $\beta\in[0.001, 0.005,0.01]$. 
Performance is shown in the estimated NMSE, with the corresponding standard error for different values of $\l$ and $\ln$. With $\l > 0.6$, the error of GQ\la increased sharply due to increased influence of importance sampling ratios. It clearly depicts the failure to perform effective multi-step learning by an importance sampling based algorithm when $\l$ is large. The error of ABQ\lna decreased substantially for most step-size combinations as $\ln$ is increased from 0 to 0.5, and it decreased slightly for some step-size combinations as $\ln$ was increased up to $1$. 
This example clearly shows that ABQ\lna can perform effective multi-step learning while avoiding importance sampling ratios. 
On the other hand, the best performance of GQ\la was better than ABQ\lna for the smallest value of the first step size (i.e., $\alpha = 0.001$). When the step size was too small, GQ\la in fact benefited from the occasional large scaling from importance sampling, whereas ABQ\lna remained conservative in its updates to be safe.

	
The Mountain Car domain is typically used for policy improvement tasks, but here we use it for off-policy policy evaluation. 
We constructed the task in such a way that the importance sampling ratios for GQ can be as large as 30, emphasizing the variance issue regarding ratios.
In this task, the car starts in the vicinity of the bottom of the valley with a small nonzero speed. The three actions are: \emph{reverse throttle}, \emph{no throttle}, and \emph{forward throttle}. Rewards are -1 at every time step, and state transitions are deterministic. The discount factor $\g$ is 0.999. The policies used in the experiments were based on a simple handcrafted policy, which chooses to move forward with full throttle if the velocity of the car was nonzero and toward the goal, and chooses to move away from the goal with full throttle otherwise.

Both the target policy and the behavior policy are based on this policy but choose to randomly explore the other actions differently. 
More specifically, when the velocity was nonzero and toward the goal, the behavior policy probabilities for the actions \emph{reverse throttle}, \emph{no throttle}, and \emph{forward throttle} are $\sb{\frac{1}{300}, \frac{1}{300}, \frac{298}{300}}$, and the target policy probabilities are $\sb{0.1, 0.1, 0.8}$, respectively. In the other case, the behavior and the target policy probabilities are $\sb{\frac{298}{300}, \frac{1}{300}, \frac{1}{300}}$ and $[0.8, 0.1, 0.1]$, respectively. We set the policies this way so that episodes complete under both policies in a reasonable number of time steps, while the importance sampling ratio may occasionally be as large as $0.1\times300=30$.

The feature vector for each state-action pair contained 32 features for each action, where only the features corresponding to the given action were nonzero, produced using tile coding with ten 4$\times$4 tilings. 
Each algorithm ran on the same 100 independent sequences of samples, each consisting 10,000 episodes. 
The performance was measured in terms of the Mean-Squared-Error (MSE) with respect to the estimated value $\hat{\q}\in\Real^{30}$ of 30 chosen state-action pairs. These pairs were chosen by running the agent under the behavior policy for 1 million time steps, restarting episodes each time termination occurs, and choosing 30 pairs uniformly randomly from the last half million time steps. 
The ground truth values for these 30 pairs $\hat{\q}$ were estimated by following the target policy 100 times from those pairs and forming the average.
The mean-squared error from $\hat{\q}$ was normalized by $\Verb{\hat{\q}}^2_{2}$. 

We use the mountain car off-policy task to illustrate through learning curves how $\l$ and $\ln$ affect GQ\la and ABQ\lna, respectively.
Figure \ref{fig:two-state-and-mountaincar} (right) shows the learning curves of ABQ\lna and GQ\la with respect to mean squared errors  for three difference values of $\l$ and $\ln$: 0, 0.4, and 0.8, and a particular step-size combination: $\a=0.1/($ ~ $\text{\# of active features})$ and $\beta = 0.0$.
These learning curves are averages over 100 independent runs. The standard errors of ABQ's estimated MSE here are smaller than the width of the curves shown. ABQ achieved a significantly lower MSE with $\ln>0$ than with $\ln=0$. On the other hand, GQ performed unreliably with larger $\l$. With $\l=0.4$, the learning curves are highly varying from each other due to occasionally having the largest ratio value 30, affecting the update at different time steps. Some learning curves even became unstable, causing the MSE to move away further and further with time, and affecting the average MSE, which explains the spikes. When $\l=0.8$ was chosen, all learning curves became unstable in few steps.


\begin{figure}[t]
 \begin{center}
 \vspace{-10pt}
\includegraphics[scale=1.1]{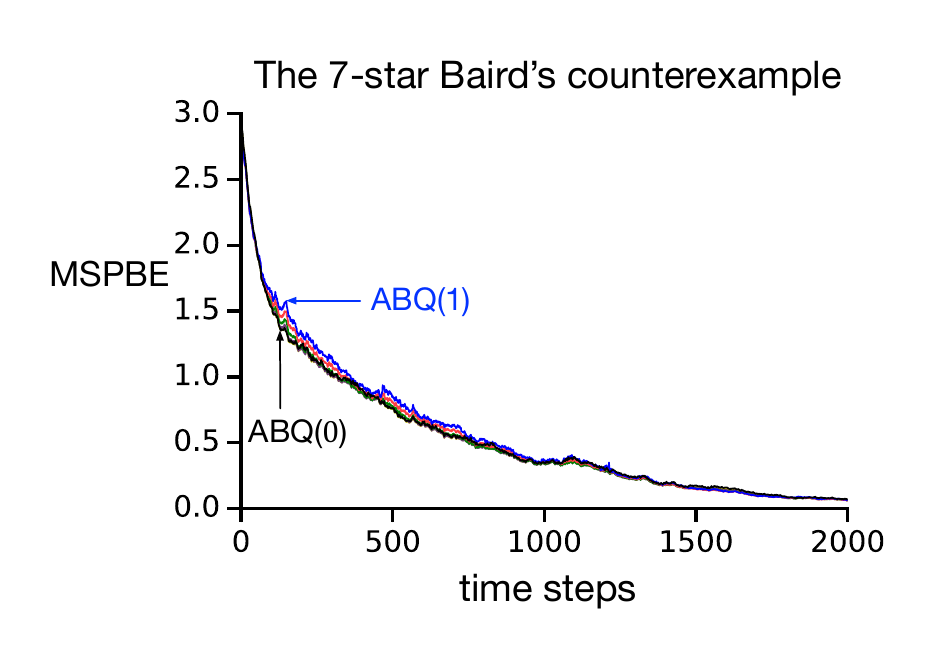}
\caption{
ABQ\lna is stable on Baird's counterexample for different values of $\ln$.
}
\label{fig:baird}
 \end{center}
\end{figure}

In the 7-star Baird's counterexample, adopted from White (\hr{2015}), step sizes were set as $\alpha=0.05$ and $\beta = 0.1$, and the bootstrapping parameter $\ln$ was chosen evenly between $0$ and $1$. The Mean Squared Projected Bellman Error (MSPBE) was estimated by averaging over 50 runs. As shown in Figure \ref{fig:baird}, ABQ\lna performed stably with all values of $\ln$ used. This validates empirically the gradient correction in ABQ\lna. 

\section{Action-dependent bootstrapping as a framework for off-policy algorithms}

ABQ\lna is a result of this new idea of varying the bootstrapping parameter in an action-dependent manner so that an explicit presence of importance sampling ratios are mitigated from the update. However, it is not the only action-dependent bootstrapping scheme one can devise. It is possible to bound the product $\l_t^n \rho_t^n$ by using other action-dependent bootstrapping schemes. Different schemes not only allow us to derive new algorithms, they may also be used to understand some existing algorithms better. Here, we show how the Retrace algorithm by Munos et al.\ (2016) can be understood in terms of a particular way of setting the action-dependent bootstrapping parameter.

Retrace is a tabular off-policy algorithm that approaches the variance issue by truncating the importance sampling ratio. We show that such truncations can be understood as varying the action-dependent bootstrapping parameter in a particular manner, first, by constructing a forward-view update with a different action-dependent bootstrapping scheme than ABQ's, second, by  deriving the asymptotic solution corresponding to that forward-view update, and third, by showing that the equivalent backward-view update is the same as the Retrace algorithm. For generality, we take these steps in the linear function approximation case and incorporate gradient corrections for stability. The resulting algorithm, we call \emph{AB-Trace\lna} is a stable generalization of Retrace.

We construct a new forward-view update similar to (\ref{eqn:abq-for}) with $\l_\zeta=\nu_\zeta\paran{s,a}\mu(a|s)$, where $\nu$ is redefined as $\nu_\zeta\paran{s,a} \deq \zeta\min\paran{\frac{1}{\pi(a|s)}, \frac{1}{\mu(a|s)}}$. Here, we treat $1/0=\infty$ and $0\cdot\infty=0$. Then we can directly use Appendix A.1 to derive its asymptotic solution:
\eqn{
\w^\zeta_\infty &\deq \A_\zeta^{-1}\b_\zeta, \\
\A_\ln	&\deq \X\tr \D_\mu \paran{\I - \g \P_\pi  \L_\zeta }^{-1} \paran{\I- \g \P_\pi } \X, \\
\b_\ln	&\deq \X\tr \D_\mu \paran{\I - \g \P_\pi \L_\zeta }^{-1} \r,
}
where the diagonal elements of $\L_\zeta$ are $\l_\zeta(s,a)=\nu_\zeta\paran{s,a}\mu(a|s)=\zeta\min\paran{\frac{1}{\pi(a|s)}, \frac{1}{\mu(a|s)}}\mu(a|s)$ for different state-action pairs.

A stable forward-view update with gradient correction corresponding to the above asymptotic solution can be derived by directly using the results of Appendix A.3.
The resulting algorithmn, AB-Trace\lna, is given by the following updates:
\eqn{
\delta_t	&\deq \Rtp + \gamma\w_t\tr\bxtp - \w_t\tr\x_t, &\\
\nu_\zeta\paran{s,a} &\deq \zeta\min\paran{\frac{1}{\pi(a|s)}, \frac{1}{\mu(a|s)}}, \\
\tilde{\x}_{t+1}	&\deq \sum_a \nu_\zeta(S_{t+1}, a) \pi(a|S_{t+1}) \x(S_{t+1}, a), &\\
\e_t		&\deq  \g\nu_{\zeta,t}\pi_t\e_{t-1} + \x_t,  \\
\w_{t+1} 	&\deq \w_t + \a_t \paran{ \delta_t \e_t - \g \e_t \tr \h_t \paran{\bxtp - \tilde{\x}_{t+1} }  } ,\\
\h_{t+1} 	&\deq \h_t + \beta_t \paran{ \delta_t \e_t - \h_t \tr\x_t\x_t  },
}
Note that, the factor $\nu_{\zeta,t}\pi_t$ in the eligibility trace vector update can be rewritten as:
\eqn{
\nu_{\zeta,t}\pi_t &= \zeta\min\paran{\frac{1}{\pi_{t}}, \frac{1}{\mu_t}}\pi_t \\
&= \zeta \min\paran{1, \rho_t}.
}
From here, it is easy to see that, if the feature representation is tabular and the gradient correction term is dropped, then the AB-Trace algorithm reduces to the Retrace algorithm.

Finally, we remark that the action-dependent bootstrapping framework provides a principled way of developing stable and efficient off-policy algorithms as well as unifying the existing ones, where AB-Trace and its connection to Retrace is only one instance.

\section{Related works}
A closely related algorithm is Tree Backup by Precup et al.\ (2000). This algorithm can also be produced as a special case of ABQ\lna, if we remove gradient correction, consider the feature vectors always to be the standard basis, and $\nu_\zeta$ to be always set to a constant, instead of setting it in an action-dependent manner. 
In the on-policy case, the Tree Backup algorithm fails to achieve the TD(1) solution, whereas ABQ achieves it with $\zeta=1$.

Our work is not a trivial generalization of this prior work.
The Tree Backup algorithm was developed using a different intuition based on backup diagrams and was introduced only for the lookup table case. 
Not only does ABQ\lna extend the Tree Backup algorithm, but the idea of action-dependent bootstrapping also played a crucial role in deriving the ABQ\lna algorithm with gradient correction in a principled way.
Another related algorithm is Retrace, which we have already shown to be a special case of the AB-Trace algorithm. The main differences are that Retrace was introduced and analyzed in the case of tabular representation, and thus Retrace is neither stable nor shown to achieve multi-step solutions in the case of function approximation. 

Yet another class of related algorithms are where the bootstrapping parameter is adapted based on past data, for example, the works by White and White (2016b) and Mann et al.\ (2016). Beside adapting on a state-action-pair basis, the main difference between those algorithms and the ones introduced here under the action-dependent bootstrapping framework is that our algorithms adapt the bootstrapping parameter using only the knowledge of policy probabilities for the current state-action pair whereas the algorithms in the other class involve a separate learning procedure for adapting $\l$ and hence are more complex.

\section{Discussion and conclusions}

In this paper, we have introduced the first model-free off-policy  algorithm ABQ\lna that can produce multi-step function approximation solutions without requiring to use importance sampling ratios. The key to this algorithm is allowing the amount of bootstrapping to vary in an action-dependent manner, instead of keeping them constant or varying only with states. Part of this action-dependent bootstrapping factor mitigates the importance sampling ratios while the rest of the factor is spent achieving multi-step bootstrapping. 
The resulting effect is that the large variance issue with importance sampling ratios is readily removed without giving up multi-step learning. This makes ABQ\lna more suitable for practical use.
Action-dependent bootstrapping provides an insightful and well-founded framework for deriving off-policy algorithms with reduced variance. The same idea may be applied to state-value estimation. The ratios cannot be eliminated completely in this case; nevertheless, reduction in the variance can be expected. According to an alternative explanation based on backup diagrams (Precup et al.\ 2000), ABQ updates may be seen as devoid of importance sampling ratios, while the action-dependent bootstrapping framework can now provide us a clearer mechanistic view of how updates without importance sampling ratios can perform off-policy learning.

The convergence of this algorithm can be analyzed similarly to the work by Karmakar and Bhatnagar (\hr{2015}) for diminishing step sizes and Yu (\hr{2015}) for constant step sizes, by using properties of the eligibility traces and stochastic approximation theory.
Investigating the possibility to include true online updates (\hr{van Seijen et al.\ 2016}) is also an interesting direction for future work.

\section*{References}

\newcommand{\hangin}{
\noindent
\hangindent=.30cm
}
{


\hangin
Dann, C., Neumann, G., Peters, J. (2014). Policy evaluation with
temporal differences: a survey and comparison. \emph{Journal of
Machine Learning Research, 15}:809--883.

\hangin
Baird, L.~C. (1995). Residual algorithms: Reinforcement learning with function approximation. In \emph{Proceedings of the 12th International Conference on Machine Learning}, pp. 30--37.

\hangin
Dud\'{i}k, M., Langford, J., Li, L. (2011). Doubly
robust policy evaluation and learning. In \emph{Proceedings of
the 28th International Conference on Machine Learning}, pp. 1097--1104.

\hangin
Geist, M., Scherrer, B. (2014). Off-policy learning with eligibility traces: A survey. \emph{Journal of Machine Learning Research, 15}:289--333.

\hangin
Hallak, A., Tamar, A., Munos, R., Mannor, S. (2015a). Generalized emphatic temporal difference learning: bias-variance analysis. \emph{arXiv preprint} arXiv:1509.05172.

\hangin
Hallak, A., Schnitzler, F., Mann, T., Mannor, S. (2015b). Off-policy model-based learning under unknown factored dynamics. In \emph{Proceedings of the 32nd International Conference on Machine Learning}, pp. 711--719.

\hangin
Hammersley, J. M. Handscomb, D. C. (1964). Monte Carlo methods, \emph{Methuen \& co. Ltd.}, London, pp. 40, 

\hangin
Harutyunyan, A., Bellemare, M. G., Stepleton, T., Munos, R. (2016). Q ($\lambda $) with off-policy corrections. \emph{arXiv preprint} arXiv:1602.04951.

\hangin
Karmakar, P., Bhatnagar, S. (2015). \!Two timescale stochastic approximation with controlled Markov noise and off-policy temporal difference learning. \!\!\emph{arXiv preprint} arXiv:1503.\\09105.

\hangin
Jiang, N., Li, L. (2015). Doubly robust off-policy evaluation for reinforcement learning. \emph{arXiv preprint} arXiv:1511.03722.


\hangin
Koller, D., Friedman, N. (2009). \emph{Probabilistic Graphical Models: Principles and Techniques}. MIT Press, 2009.

\hangin
Li, L., Munos, R., Szepesvari, C. (2015). Toward 
minimax off-policy value estimation. In \emph{Proceedings
of the Eighteenth International Conference on Artificial
Intelligence and Statistics}, pp. 608--616.

\hangin
Liu, J.~S. (2001). \emph{Monte Carlo strategies in scientific computing}. Berlin, Springer-Verlag.

\hangin
Maei, H.~R., Sutton, R.~S. (2010). 
GQ\la: A general gradient algorithm for temporal-difference prediction learning with eligibility traces. 
In \emph{Proceedings of the Third Conference on Artificial General Intelligence}, pp.~91--96. Atlantis Press.

\hangin
Maei, H.~R. (2011).
\emph{Gradient Temporal-Difference Learning Algorithms}. 
PhD thesis, University of Alberta. 

\hangin
Mahmood, A.~R., van Hasselt, H., Sutton, R.~S. (2014). Weighted importance sampling for off-policy learning with linear function approximation. In \emph{Advances in Neural Information Processing Systems 27}, Montreal, Canada.

\hangin
Mahmood, A.~R., Sutton, R.~S. (2015). Off-policy learning based on weighted importance sampling with linear computational complexity. In \emph{Proceedings of the
31st Conference on Uncertainty in Artificial Intelligence}.

\hangin
Mahmood, A.~R., Yu, H., White, M., Sutton, R.~S. (2015). Emphatic temporal-difference learning. \emph{European Workshop on Reinforcement Learning 12},\emph{arXiv preprint}  ArXiv:1507.\\01569.

\hangin
Mann, T.~A., Penedones, H., Mannor, S., Hester, T. (2016). Adaptive lambda least-squares temporal difference learning. {arXiv preprint} arXiv:1612.09465.

\hangin
Paduraru, C. (2013). \emph{Off-policy Evaluation in Markov Decision Processes}, PhD thesis, McGill University.

\hangin
Precup, D., Sutton, R.~S., Singh, S. (2000).
Eligibility traces for off-policy policy evaluation. 
In \emph{Proceedings of the 17th International Conference on Machine Learning}, 
pp.~759--766. Morgan Kaufmann.

\hangin
Precup, D., Sutton, R.~S., Dasgupta, S. (2001). Off-policy temporal-difference learning with function approximation. In \emph{Proceedings of the 18th International Conference on Machine Learning}.


\hangin
Munos, R, Stepleton, T, Harutyunyan, A, Bellemare, M.~G. (2016).
Safe and efficient off-policy reinforcement learning.
In \emph{Proceedings of Neural Information Processing Systems}. 

\hangin
Rubinstein, R.~Y. (1981). \emph{Simulation and the Monte Carlo Method}, New York, Wiley.

\hangin
Sutton, R.S., Singh, S.P. (1994). On bias and step size in temporal-difference learning. In \emph{Proceedings of the Eighth Yale Workshop on Adaptive and Learning Systems}, pp. 91-96.

\hangin
Sutton, R.~S., Barto, A.~G. (1998). \emph{Reinforcement Learning: An Introduction}. MIT Press.

\hangin
Sutton, R. S., Precup, D., Singh, S. (1999). Between MDPs and semi-MDPs: A framework for temporal abstraction in reinforcement learning. \emph{Artificial intelligence}, 112(1), 181--211.


\hangin
Sutton, R. S., Mahmood, A. R., Precup, D., van Hasselt, H. (2014). A new Q\la with interim
forward view and Monte Carlo equivalence. In \emph{Proceedings of the 31st International
Conference on Machine Learning}. JMLR W\&CP 32(2).

\hangin
Sutton, R. S., Mahmood, A. R., White, M. (2016). An emphatic approach to the problem of off-policy temporal-difference learning. \emph{Journal of Machine Learning Research 17}, (73):1--29.

\hangin
Thomas, P. S. (2015). \emph{Safe Reinforcement Learning}. PhD thesis,
University of Massachusetts Amherst.

\hangin
Thomas, P. S., Brunskill, E. (2016). Data-efficient off-policy policy evaluation for reinforcement learning. \emph{arXiv preprint} arXiv:1604.00923.

\hangin
Tsitsiklis, J. N., Van Roy, B. (1997). An analysis of temporal-difference learning with function approximation. \emph{IEEE Transactions on Automatic Control}, 42(5), 674--690.

\hangin
van Hasselt, H. (2011). \emph{Insights in Reinforcement Learning: formal analysis and empirical evaluation of temporal-difference learning algorithms}. PhD thesis, Universiteit Utrecht.

\hangin
van Hasselt, H., Mahmood, A.~R., Sutton, R.~S. (2014). Off-policy TD\la with a true online equivalence. In \emph{Proceedings of the 30th Conference on Uncertainty in Artificial Intelligence}, Quebec City, Canada.

\hangin
van Seijen, H., \& Sutton, R.~S. (2014).
True online TD($\l$).
In \emph{Proceedings of the 31st International Conference on Machine Learning}.
JMLR W\&CP 32(1):692--700.

\hangin
van Seijen, H., Mahmood, A.~R., Pilarski, P.~M., Machado, M.~C., Sutton, R.~S. (2016). 
True online temporal-difference learning. 
\emph{Journal of Machine Learning Research 17}
(145):1--40.

\hangin
White, A., Modayil, J., Sutton, R.~S. (2012). Scaling life-long off-policy learning. In \emph{Proceedings of the Second Joint IEEE International Conference on Development and Learning and on Epigenetic Robotics}, San Diego, USA.

\hangin
White, A. (2015). \emph{Developing a Predictive Approach to Knowledge}. PhD thesis, University of Alberta.

\hangin
White, A., White, M. (2016a). Investigating practical, linear temporal difference learning. In \emph{Proceedings of the 2016 International Conference on Autonomous Agents \& Multiagent Systems}, 494--502.

\hangin
White, M., White, A. (2016b). A greedy approach to adapting the trace parameter for temporal difference learning. In \emph{Proceedings of the 2016 International Conference on Autonomous Agents \& Multiagent Systems}, 557--565.

\hangin
Yu, H. (2012). Least squares temporal difference methods: An analysis under general conditions. \emph{SIAM Journal on Control and Optimization, 50}(6), 3310--3343.

\hangin
Yu, H. (2015). On convergence of emphatic temporal-difference learning. \emph{arXiv preprint} arXiv:1506.02582; a shorter version appeared in {The 28th Annual Conference
on Learning Theory (COLT) 2015}.

\hangin
Yu, H. (2016). Weak convergence properties of constrained emphatic temporal-difference learning with constant and slowly diminishing stepsize. \emph{Journal of Machine Learning Research  17}(220):1--58.

}

\onecolumn
\newpage

\section*{Appendix}

\section*{A.1 Derivation of the ABQ\lna solution}

We follow the approach taken by Sutton et al.\ (2016) to derive the asymptotic solution. The key-step in their approach is to derive the expected update corresponding to the stochastic update. 

The off-line backward-view update defined by (\ref{eqn:abq-for}) can be rewritten as:
\eqn{
\Delta\w_t &= \alpha_t \paran{ H_t^{\zeta} - \w^\top \x_t }\x_t \\
&= \alpha_t \sum_{n=t}^\infty  \g^{n-t} \nu_{\zeta,t+1}^{n} \pi_{t+1}^{n} \delta_{n}  \x_t \\
&= \alpha_t \sum_{n=t}^\infty  \g^{n-t} \nu_{\zeta,t+1}^{n} \pi_{t+1}^{n} \paran{ R_{n+1} + \gamma\w\tr\bx_{n+1} - \w\tr\x_n } \x_t \\
&= \alpha_t \paran{  \underbrace{\sum_{n=t}^\infty  \g^{n-t} \nu_{\zeta,t+1}^{n} \pi_{t+1}^{n}  R_{n+1} \x_t}_{\b_t} - \underbrace{\sum_{n=t}^\infty  \g^{n-t} \nu_{\zeta,t+1}^{n} \pi_{t+1}^{n}\x_t \paran{\x_n - \g\bx_{n+1}}\tr}_{\A_t}\w } \\
&= \alpha_t \paran{ \b_t - \A_t \w }.
}

With $\w$ held fixed, the expected update direction for (\ref{eqn:abq-for}) is given by the expectation of $\b_t - \A_t \w$ with respect to the stationary Markov chain $\{(S_t, A_t, R_{t+1})\}$ induced by $\mu$. 
Specifically, let $\E_0$ denote expectation with respect to this stationary Markov chain.  
We calculate the expectation $\E_0 \sb{\A_t}$ (for any fixed $t$) as:
\eqn{
\E_0\sb{\A_t}&= \E_0\sb{ \sum_{n=t}^\infty  \g^{n-t} \nu_{\zeta,t+1}^{n} \pi_{t+1}^{n}\x_t \paran{\x_n - \g\bx_{n+1}}\tr } \\
&= \sum_{s,a} d_\mu(s,a) \E_\mu\sb{ \sum_{n=t}^\infty  \g^{n-t} \nu_{\zeta,t+1}^{n} \pi_{t+1}^{n}\x_t \paran{\x_n - \g\bx_{n+1}}\tr \Big| S_t=s, A_t=a } \\
&= \sum_{s,a} \x(s,a) d_\mu(s,a)  \underbrace { \E_\mu\sb{  \sum_{n=t}^\infty \g^{n-t} \nu_{\zeta,t+1}^{n} \pi_{t+1}^{n} \paran{\x_n - \g\bx_{n+1}}\tr \Big| S_t=s, A_t=a } }_{\e\tr(s,a)\in\Real^n} \\
&= \sum_{s,a} \x(s,a) d_\mu(s,a) \e(s,a)\tr \\
&= \X\tr \D_\mu \EM.
}
Here the matrix $\EM\in\Real^{|\St|\cdot|\Act|\times n}$ is such that $[\EM]_{sa,:} \deq \e(s,a)$. We can write $\e(s,a)\tr$ recursively as:
\eqn{
\e(s,a)\tr &= \E_\mu\sb{  \sum_{n=t}^\infty \g^{n-t} \nu_{\zeta,t+1}^{n} \pi_{t+1}^{n} \paran{\x_n - \g\bx_{n+1}}\tr \Big| S_t=s, A_t=a }  \\
&= \E_\mu\sb{ \paran{\x_t - \g\bx_{t+1}}\tr \Big| S_t=s, A_t=a } \\
&+ \E_\mu\sb{  \sum_{n=t+1}^\infty \g^{n-t} \l_{\zeta,t+1}^{n} \rho_{t+1}^{n} \paran{\x_n - \g\bx_{n+1}}\tr \Big| S_t=s, A_t=a } \\
&=  \paran{ \x(s,a)\tr - \E_\mu\sb{\bx_{t+1}\tr \Big| S_t=s, A_t=a } } \\
&+  \g  \E_\mu\sb{ \l_{\zeta, t+1} \rho_{t+1}  \sum_{n=t+1}^\infty \g^{n-t-1} \l_{\zeta, t+2}^{n} \rho_{t+2}^{n} \paran{\x_n - \g\bx_{n+1}}\tr \Big| S_t=s, A_t=a } \\
&= \paran{ \x(s,a)\tr - \g \sum_{s'} p(s'|s,a) \sum_{a'} \pi(a'|s') \x(s',a')\tr}\\
&+  \g \sum_{s'a'} p(s'|s,a) \mu(a'|s')  \l_\zeta(s',a')\frac{\pi(a'|s')}{\mu(a'|s')}\\
&\times \E_\mu\sb{ \sum_{n=t+1}^\infty \g^{n-t-1} \nu_{\zeta, t+2}^{n} \pi_{t+2}^{n} \paran{\x_n - \g\bx_{n+1}}\tr \Big| S_{t+1}=s', A_{t+1}=a' } \\
&= \paran{ \x(s,a)\tr - \g \sum_{s',a'}[\P_\pi]_{sa,s'a'} \x(s',a')\tr}\\
&+  \g \sum_{s'a'} p_\pi(s',a'|s,a) \l_\zeta(s',a') \e(s',a')\tr\\
&= \paran{ \x(s,a)\tr - \g \sum_{s',a'}[\P_\pi]_{sa,s'a'} \x(s',a')\tr} +  \g \sum_{s'a'} [\P_\pi \L_{\zeta}]_{sa,s'a'} \e(s',a')\tr.
}
Therefore, we can write $\EM$ recursively as:
\eqn{
\EM &= (\I - \g \P_\pi )\X + \g \P_\pi \L_{\zeta} \EM \\
&= (\I - \g \P_\pi )\X + \g \P_\pi \L_{\zeta}(\I - \g \P_\pi )\X +  \paran{ \g \P_\pi \L_{\zeta}}^2(\I - \g \P_\pi )\X + \cdots \\
&= \paran{\I + \g \P_\pi \L_{\zeta} +  \paran{\g \P_\pi \L_{\zeta}}^2+\cdots}(\I - \g \P_\pi )\X \\
&= (\I - \g \P_\pi \L_{\zeta})^{-1} (\I - \g \P_\pi )\X.
}
It then follows that
\eqn{
\E_0\sb{\A_t}&= \X\tr \D_\mu \EM = \X\tr \D_\mu (\I - \g \P_\pi \L_{\zeta})^{-1} (\I - \g \P_\pi )\X \\
&= \A_\zeta.
}
The proof of $\E_0[ \b_t ] = \b_\zeta$ is similar, and we omit it here.

Thus we have shown that the expected update corresponding to (\ref{eqn:abq-for}) is:
\eqn{
\Delta\w = \E_0[\Delta\w_t] = \a (\b_\zeta - \A_\zeta \w ).
}
When $\A_\zeta$ is invertible, $\w^\zeta_\infty = \A_\zeta^{-1} \b_\zeta$ is the desired solution that (\ref{eqn:abq-for}) aims to attain in the limit. 
Note, however, that this expected update $\Delta\w = \a (\b_\zeta - \A_\zeta \w )$ may not be stable, which is a common problem with many off-policy temporal-difference learning algorithms. A modification to (\ref{eqn:abq-for}), such as gradient correction, is needed to ensure stability, which is attained by the ABQ\lna algorithm described in Section 6. 

~

~

~

~

~

\section*{A.2 Derivation of the backward-view update of (\ref{eqn:abq-for})}

Using the forward-view update given by (\ref{eqn:abq-for}), the total update can be given by:
\eqn{
\sum_{t=0}^\infty \Delta\w_t &= \sum_{t=0}^\infty \alpha_t \lt( H^{\zeta}_t - \w\tr\x_t \rt) \x_t \\
&= \sum_{t=0}^\infty \sum_{n=t}^\infty  \g^{n-t} \nu_{\ln,t+1}^{n} \pi_{t+1}^{n} \delta_{n} \x_t \\
&= \sum_{t=0}^\infty\a_t \sum_{n=0}^t \g^{t-n} \nu_{\ln,n+1}^{t}  \pi_{n+1}^{t} \delta_{t} \x_n \\
&= \sum_{t=0}^\infty\a_t \delta_{t}  \underbrace{\sum_{n=0}^t \g^{t-n} \nu_{\ln,n+1}^{t}  \pi_{n+1}^{t} \x_n}_{\e_t} \\
&= \sum_{t=0}^\infty \a_t\delta_{t} \e_t.
}
Therefore, the backward-view update can be written as:
\eqn{
\Delta\w_t^B = \a_t\delta_{t} \e_t.
}
The eligibility trace vector $\e_t\in\Real^n$ can be written recursively as:
\eqn{
\e_t 	&= \sum_{n=0}^t \g^{t-n} \nu_{\ln,n+1}^{t}  \pi_{n+1}^{t} \x_n \\
&= \g \nu_{\ln,t} \pi_t \sum_{n=0}^{t-1} \g^{t-n-1} \nu_{\ln,n+1}^{t-1}  \pi_{n+1}^{t-1} \x_n + \x_t \\
&= \g \nu_{\ln,t} \pi_t \e_{t-1} + \x_t.
}

\newpage

\section*{A.3 Derivation of ABQ\lna}

The key step in deriving a gradient-based TD algorithm, as proposed by Maei (\hr{2011}), is to formulate an associated Mean Squared Projected Bellman Error (MSPBE) and produce its gradient, which can then be sampled to produce stochastic updates.

The MSPBE for the update (\ref{eqn:abq-for}) is given by:
\eqn{
J(\w)		&= \Verb{ \Pi_\mu T_\pi^{(\L_{\zeta})}\X\w - \X\w  }_{\D_\mu}^2 \\
&= \Verb{ \Pi_\mu \paran{ T_\pi^{(\L_{\zeta})}\X\w -  \X\w }  }_{\D_\mu}^2 \\
&= \paran{ T_\pi^{(\L_{\zeta})}\X\w - \X\w}\tr \Pi_\mu\tr \D_\mu \Pi_\mu \paran{ T_\pi^{(\L_{\zeta})}\X\w - \X\w} \\
&= \paran{ \X\tr\D_\mu\paran{T_\pi^{(\L_{\zeta})} \X\w - \X\w}}\tr  (\X\tr \D_\mu \X)^{-1} \X\tr \D_\mu\paran{T_\pi^{(\L_{\zeta})}  \X\w - \X\w}.
}
Here, the Bellman operator corresponding to the bootstrapping matrix $\L_\zeta$ is defined for all $\q\!\in\!\Real^{|\St|\cdot|\Act|}$ as 
\[T_\pi^{(\L_{\zeta})} \q \deq \paran{\I - \g\P_\pi\L_{\zeta}}^{-1} \sb{\r + \g\P_\pi(\I-\L_{\zeta}) \q}.\]
\vspace{-10pt}
\eqnl{
\text{Let~}\gb&\deq \X\tr\D_\mu\paran{T_\pi^{(\L_{\zeta})} \X\w -  \X\w} \\&= 
\X\tr\D_\mu\paran{ \paran{\I - \g\P_\pi\L_{\zeta}}^{-1} \sb{\r + \g\P_\pi(\I-\L_{\zeta}) \X\w } - \X\w} & \\
&=\X\tr\D_\mu\paran{ \paran{\I - \g\P_\pi\L_{\zeta}}^{-1} \sb{\r + \paran{ (\I-\g\P_\pi\L_{\zeta}) - (\I-\g\P_\pi) } \X\w } - \X\w} & \\
&=\X\tr\D_\mu\paran{ {  \paran{\I - \g\P_\pi\L_{\zeta}}^{-1}\r + \X\w -  \paran{\I - \g\P_\pi\L_{\zeta}}^{-1}(\I-\g\P_\pi) } \X\w  - \X\w} & \\
&=\X\tr\D_\mu{ {  \paran{\I - \g\P_\pi\L_{\zeta}}^{-1}\paran{\r - (\I-\g\P_\pi) \X\w } } }.& 
}
Also let $\C \deq (\X\tr \D_\mu \X)$. Then the gradient can be written as:
\eqn{
\nabla J(\w)		&= -\half\paran{\X\tr\D_\mu{ {  \paran{\I - \g\P_\pi\L_{\zeta}}^{-1}\paran{ \I-\g\P_\pi} \X  } }}\tr \C^{-1} \gb \\
&= -\half\paran{\X\tr\D_\mu{ {  \paran{\I - \g\P_\pi\L_{\zeta}}^{-1}\paran{ \paran{\I - \g\P_\pi\L_{\zeta}}\X - \g\P_\pi(\I-\L_{\zeta})\X  } } }}\tr \C^{-1} \gb \\
&= - \half\paran{\X\tr\D_\mu  \X - \g \X\tr\D_\mu \paran{\I - \g\P_\pi\L_{\zeta}}^{-1} \P_\pi(\I-\L_{\zeta})\X  }  \tr \C^{-1} \gb \\ 
&= - \half\paran{ \gb - \g \paran{ \underbrace{\X\tr\D_\mu\paran{\I - \g  \P_\pi\L_{\zeta}}^{-1} \P_\pi(\I-\L_{\zeta})\X}_{\H}}\tr \C^{-1}\gb } \\
&= - \half\paran{ \gb - \g \H\tr \C^{-1}\gb }.
}
So if we know the gradient, the gradient-descent step would be to add the following to the parameter vector:
\eqn{
 -\half \a_t \nabla J(\w)  &=  \a_t \paran{ \gb - \g \H\tr \C^{-1}\gb }.
}

Now to derive the stochastic updates for ABQ\lna, let us consider a double-ended \emph{stationary} Markov chain induced by $\mu$, $\{\ldots, (S_{-2}, A_{-2}, R_{-1}), (S_{-1}, A_{-1}, R_0), (S_0, A_0, R_1),  (S_1, A_1,$ $R_2), \ldots\}$. Let $\E_0$ denote expectation with respect to the probability distribution of this stationary Markov chain. Fix $t$ to be any integer. Then we can write the first term $\gb$ in $\nabla J(\w)$ in expectation form as follows:
\eqn{
\gb 	&= \X\tr\D_\mu{ {  \paran{\I - \g\P_\pi\L_{\zeta}}^{-1}\paran{\r - (\I-\g\P_\pi) \X\w } } } \\
	&= \X\tr\D_\mu \paran{\I - \g\P_\pi\L_{\zeta}}^{-1}\r - \X\tr\D_\mu \paran{\I - \g\P_\pi\L_{\zeta}}^{-1} (\I-\g\P_\pi) \X\w \\
	&= \b_\zeta - \A_\zeta\w; ~~~~~ \text{($\A_\zeta$ and $\b_\zeta$ as defined by (\ref{eqn:Aabq}) and (\ref{eqn:babq}) )} \\
	&= \E_0\sb{ \paran{ H_t^{\zeta} - \w^\top \x_t }\x_t} \\
	&= \E_0\sb{ \sum_{n=t}^\infty  \g^{n-t} \nu_{\zeta,t+1}^{n} \pi_{t+1}^{n} \delta_{n} \x_t } 	 \\
	&= \E_0\sb{ \delta_t \x_t +  \sum_{n=t+1}^\infty  \g^{n-t} \nu_{\zeta,t+1}^{n} \pi_{t+1}^{n} \delta_{n} \x_t }\\
	&= \E_0\sb{ \delta_t \x_t + \sum_{n=t}^\infty  \g^{n-(t-1)} \nu_{\zeta,t}^{n} \pi_{t}^{n} \delta_{n} \x_{t-1} }; ~~~\text{shifting indices and using stationarity}\\
	&= \E_0\sb{ \delta_t \x_t + \g\nu_{\zeta,t}\pi_t\sum_{n=t}^\infty  \g^{n-t} \nu_{\zeta,t+1}^{n} \pi_{t+1}^{n} \delta_{n} \x_{t-1} }\\
	&= \E_0\sb{ \delta_t \x_t + \g\nu_{\zeta,t}\pi_t\paran{ \delta_t \x_{t-1} + \sum_{n=t+1}^\infty  \g^{n-t} \nu_{\zeta,t+1}^{n} \pi_{t+1}^{n} \delta_{n} \x_{t-1} }  }\\
	&= \E_0\sb{ \delta_t\paran{ \x_t + \g\nu_{\zeta,t}\pi_t \x_{t-1} + \cdots }}; \qquad \text{shifting indices and using stationarity} \\
	&= \E_0\sb{ \delta_t \e_t},
}
where $\e_t$ is a well-defined random variable and can be written recursively as:
\eqn{
\e_t &= \x_t + \g\nu_{\zeta,t}\pi_t \x_{t-1} + \cdots = \x_t + \g\nu_{\zeta,t}\pi_t\e_{t-1}.
}

Similarly, we can also express the term $\H$ in $\nabla J(\w)$ in expectation form. Let us define
\eqn{
\tilde{\x}_t = \sum_{a} \l_\zeta(S_t, a) \pi(a|S_t) \x(S_t, a).
}
Then we can write:
\eqn{
\H	&= \X\tr\D_\mu\paran{\I - \g  \P_\pi\L_{\zeta}}^{-1} \P_\pi(\I-\L_\zeta)\X \\
	&= \X\tr\D_\mu \P_\pi(\I-\L_{\zeta})\X + \X\tr\D_\mu \g \P_\pi\L_{\zeta} \P_\pi(\I-\L_\zeta)\X + \cdots \\
	&= \sum_{s,a} d_\mu(s,a) \x(s,a) \sum_{s'a'} p(s'|s,a) \pi(a'|s') (1-\l_\zeta(s',a')) \x(s',a')\tr   \\
	&~+ \X\tr\D_\mu \g \P_\pi\L_{\zeta} \P_\pi(\I-\L_\zeta)\X + \cdots \\
	&= \E_0 \sb{ \x_t \paran{\bxtp - \tilde{\x}_{t+1} }\tr }  \\
	&~+ \g\sum_{s,a} d_\mu(s,a)\x(s,a) \sum_{s',a'} p(s'|s,a) \pi(a'|s')\zeta(a'|s')\mu(a'|s')  \\
	&~~\times\sum_{s''a''} p(s''|s',a') \pi(a''|s'') (1-\l_{\zeta}(s'',a'')) \x(s'',a'')\tr + \cdots \\	
	&=  \E_0 \sb{ \x_t\paran{\bxtp - \tilde{\x}_{t+1} }\tr    } \\
	&~+ \E_0 \sb{ \g \nu_{\zeta,t+1}\pi_{t+1} \x_{t} \paran{\bx_{t+2} - \tilde{\x}_{t+2} }\tr  }+\cdots;~~~\text{shifting indices and using stationarity} \\
	&=  \E_0 \sb{\x_t\paran{\bxtp - \tilde{\x}_{t+1} }\tr  } \\
	&~+ \E_0 \sb{ \g \nu_{\zeta,t}\pi_{t} \x_{t-1}\paran{\bxtp - \tilde{\x}_{t+1} }\tr  }+\cdots; ~~~\text{shifting indices and using stationarity} \\
	&= \E_0 \sb{ \paran{\x_t +  \g \nu_{\zeta,t}\pi_{t} \x_{t-1} + \cdots } \paran{\bxtp - \tilde{\x}_{t+1} }\tr   ) } \\
	&= \E_0 \sb{ \e_t \paran{\bxtp - \tilde{\x}_{t+1} }\tr    }.
}

Therefore, a stochastic update corresponding to the expected gradient-descent update can be written as:
\eqn{
\Delta \w 	&= \a_t \paran{ \delta_t \e_t - \g \paran{\bxtp - \tilde{\x}_{t+1} }  \e_t \tr \C^{-1}\gb }.
}
The vector $\C^{-1}\gb$ can be estimated from samples by LMS but at a faster time scale and with a larger step-size parameter $\beta$:
\eqn{
\Delta\h = \beta_t \paran{ \delta_t \e_t - \h\tr\x_t\x_t  }.
}
It can be shown that with $\w$ held fixed, under standard diminishing step-size rules for $\beta_t$, the $\{\h_t\}$ produced by the above updates converges to
\eqn{
\h_\infty 	&= \paran{\E_0\sb{ \x_t\x_t\tr } }^{-1} \E_0\sb{ \delta_t \e_t }  \\
		&= \paran{\X\tr \D_\mu \X}^{-1} \E_0\sb{ \delta_t \e_t }  \\
		&= \C^{-1} \gb.
}

Putting these pieces together, and also allowing $\w$ and $\h$ to vary over time, we obtain the updates of the on-line gradient-based TD algorithm, which we call ABQ\lna:
\eqn{
\w_{t+1} 	&= \w_t + \a_t \paran{ \delta_t \e_t - \g \e_t \tr \h_t \paran{\bxtp - \tilde{\x}_{t+1} }  },\\
\e_t		&=  \g\nu_{\zeta,t}\pi_t\e_{t-1} + \x_t,  \\
\h_{t+1} 	&= \h_t + \beta_t \paran{ \delta_t \e_t - \h_t \tr\x_t\x_t  }.
}

\end{document}